%% file: main.tex
\documentclass[10pt,twocolumn,letterpaper]{article}

\usepackage[pagenumbers]{cvpr} % To force page numbers, e.g. for an arXiv version
\usepackage{array}
\usepackage{multirow}


\usepackage{xcolor}

\definecolor{cvprblue}{rgb}{0.21,0.49,0.74}
\usepackage[pagebackref,breaklinks,colorlinks,allcolors=cvprblue]{hyperref}
\usepackage{float}
\usepackage{stfloats}

\def\paperID{10912} % *** Enter the Paper ID here
\def\confName{CVPR}
\def\confYear{2025}

\title{BiPO: Bidirectional Partial Occlusion Network for Text-to-Motion Synthesis}
\author{Seong-Eun Hong\\
Kyung Hee University\\
Korea, Republic\\
{\tt\small zen152@khu.ac.kr}
\and
Soobin Lim\\
Kyung Hee University\\
Korea, Republic\\
{\tt\small dnpcs@khu.ac.kr}
\and
Juyeong Hwang\\
Kyung Hee University\\
Korea, Republic\\
{\tt\small dudyyyy4@khu.ac.kr}
\and
Minwook Chang\\
NC Research, NCSOFT Corp.\\
Korea, Republic\\
{\tt\small \texttt{minwook@ncsoft.com}}
\and
Hyeongyeop Kang\\
Korea University\\
Korea, Republic\\
{\tt\small \texttt{siamiz\_hkang@korea.ac.kr}}
}

\begin{document}
% \maketitle

\twocolumn[{%
\renewcommand\twocolumn[1][]{#1}%
\maketitle
\begin{center}
    \centering
    \captionsetup{type=figure}
    \includegraphics[width=0.99\textwidth]{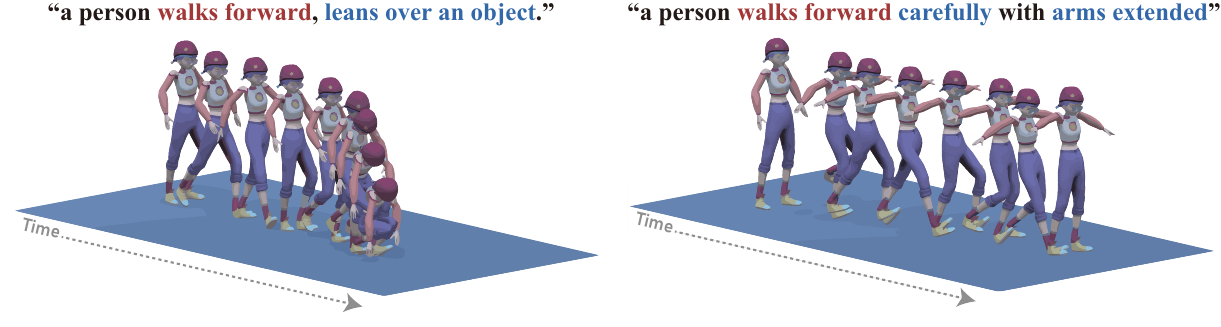}
    \captionof{figure}{BiPO generates diverse, high-quality 3D human motions from text prompts, capturing subtle nuances and motion details.}
\end{center}
}]

\input{sec/0_abstract}    
\input{sec/1_introduction}
\input{sec/2_related_work}
\input{sec/3_approach}
\input{sec/4_experiments}
\input{sec/5_user_study}
\input{sec/6_conclusion}
{
    \small
    \bibliographystyle{ieeenat_fullname}
    \bibliography{main}
}

\input{sec/X_suppl}
\end{document}

%% file: sec/0_abstract.tex
\begin{abstract}
Generating natural and expressive human motions from textual descriptions is challenging due to the complexity of coordinating full-body dynamics and capturing nuanced motion patterns over extended sequences that accurately reflect the given text.
To address this, we introduce BiPO, \textbf{Bi}directional \textbf{P}artial \textbf{O}cclusion Network for Text-to-Motion Synthesis, a novel model that enhances text-to-motion synthesis by integrating part-based generation with a bidirectional autoregressive architecture. This integration allows BiPO to consider both past and future contexts during generation while enhancing detailed control over individual body parts without requiring ground-truth motion length. 
To relax the interdependency among body parts caused by the integration, we devise the Partial Occlusion technique, which probabilistically occludes the certain motion part information during training. 
In our comprehensive experiments, BiPO achieves state-of-the-art performance on the HumanML3D dataset, outperforming recent methods such as ParCo, MoMask, and BAMM in terms of FID scores and overall motion quality. Notably, BiPO excels not only in the text-to-motion generation task but also in motion editing tasks that synthesize motion based on partially generated motion sequences and textual descriptions.  
These results reveal the BiPO's effectiveness in advancing text-to-motion synthesis and its potential for practical applications. 
\end{abstract}

%% file: sec/1_introduction.tex
\section{Introduction}
\label{sec:intro}

Text-to-motion generation is vital for applications in animation~\cite{kappel2021high}, virtual reality~\cite{guo2022generating}, video games~\cite{majoe2009enhanced, yeasin2004multiobject}, and robotics~\cite{antakli2018intelligent, koppula2013learning, koppula2015anticipating}. 
By enabling the synthesis of natural and expressive motions from textual descriptions, it bridges the gap between human language and machine-generated motion, facilitating more intuitive content creation and human-computer interaction.
However, existing methods often struggle with modeling complex full-body dynamics, leading to oversimplified representations that lack nuanced coordination between body parts.

To address this challenge, recent approaches~\cite {ghosh2021synthesis, zhong2023attt2m} treat each body part independently to capture unique motion patterns, but they often suffer from a lack of coherence between parts. ParCo~\cite{zou2024parco} addresses this by leveraging information from all parts during training, enhancing global motion coherence. Yet, its unidirectional autoregressive architecture limits the ability to anticipate future actions, hindering coordination over long temporal horizons.

Consider the description: ``\textit{The person takes side steps to his left then to his right.}" These interconnected actions require symmetrical and balanced transitions, necessitating an understanding of the overall motion pattern. A unidirectional model, unaware of the forthcoming rightward step, may produce uncoordinated movements, compromising the naturalness of the motion.

This limitation can be alleviated by bidirectional approaches, such as MoMask~\cite{guo2024momask} and BAMM~\cite{pinyoanuntapong2024bamm}, which leverage both past and future contexts for a more comprehensive understanding of the sequence.
However, these methods do not incorporate part-based generation, thereby lacking detailed control over individual body parts, and some also require ground-truth motion length, which is impractical in real-world applications.

We introduce BiPO, \textbf{Bi}directional \textbf{P}artial \textbf{O}cclusion Network for Text-to-Motion Synthesis. This is the first model integrating part-based generation with bidirectional autoregressive architecture for text-to-motion synthesis without requiring ground-truth motion length. 
This integration, however, can cause excessive interdependence among body parts, as the model tends to rely on contextual cues from neighboring parts, weakening the independence of each part's representation. 
To overcome this, we additionally devise the Partial Occlusion (PO) technique to relax the dependency among body parts during training, achieving state-of-the-art performance on HumanML3D~\cite{guo2022generating}.

In summary, our main contributions are:
\begin{itemize}
    \item We introduce BiPO, the first model to integrate part-based and bidirectional autoregressive approaches, combining the advantages of detailed control over individual body parts with global motion coherence, without the need for ground-truth motion length as input.
    \item We propose the PO, a novel training technique that mitigates excessive interdependency among body parts in bidirectional models, promoting robust and independent part representations. 
    \item Our model achieves superior performance over recent state-of-the-art models such as ParCO, MoMask, and BAMM on HumanML3D, demonstrating better FID scores and overall motion quality. 
\end{itemize}

%% file: sec/2_related_work.tex
\section{Related Work}
\label{sec:Related Work}

\paragraph{Text-to-motion generation} 
Text-to-motion generation synthesizes realistic human motions from natural language descriptions. 
Previous research has explored conditioning modalities like action classes~\cite{cervantes2022implicit, guo2020action2motion, lucas2022posegpt, petrovich2021action}, images~\cite{chen2022learning, rempe2021humor}, pose sequences~\cite{duan2021single, oreshkin2023motion, liu2022investigating, mao2019learning}, control signals~\cite{starke2022deepphase, starke2019neural, peng2021amp}, and audio~\cite{gong2023tm2d, siyao2022bailando, tseng2023edge, zhou2023ude, lee2019dancing, li2022danceformer, li2021ai}. Unconditional motion generation models have also been developed~\cite{raab2023modi, yan2019convolutional, zhang2020perpetual, zhao2020bayesian}. However, these methods often lack the expressiveness of natural language. 

To address this limitation, text-to-motion generation has gained prominence.
Early models aligned text and motion in a shared latent space~\cite{ahn2018text2action, ahuja2019language2pose} but struggled with temporal dependencies. Variational AutoEncoder (VAE)- and transformer-based models~\cite{radford2021learning, guo2022generating, tevet2022motionclip, petrovich2022temos, athanasiou2022teach, petrovich2023tmr} facilitate latent representation learning but often produce less detailed motions due to limitations in modeling complex data distributions~\cite{bredell2023explicitly}. Generative Adversarial Networks (GANs)-based models~\cite{cai2018deep, wang2020learning, lin2018human} offer sharper motions and more detailed motions but suffer from training instability and mode collapse~\cite{chen2023executing, arjovsky2017towards, salimans2016improved}, reducing diversity.

Diffusion-based models~\cite{tevet2023human, chen2023executing, zhang2022motiondiffuse, kim2023flame, zhang2023remodiffuse, gao2024guess, sun2024lgtm} also have achieved high-quality motion generation, effectively modeling complex data distributions to capture realistic, temporally coherent motions. However, their iterative denoising process~\cite{chen2023executing} demands substantial computational resources. Additionally, they often necessitate ground-truth motion lengths, restricting flexibility for variable-length sequences.

An alternative line of research utilizes Vector Quantized-Variational AutoEncoders (VQ-VAE)~\cite{van2017neural} to tokenize motions, enabling discrete representation learning. Based on the generation framework, these approaches can be categorized into unidirectional autoregressive methods and bidirectional methods.

Unidirectional autoregressive methods~\cite{zhang2023generating, zhong2023attt2m, jiang2023motiongpt, zou2024parco} generate motions sequentially but may accumulate errors over time and struggle with capturing long-term dependencies.
Bidirectional methods~\cite{guo2024momask, pinyoanuntapong2024mmm} consider both past and future contexts, leading to higher-quality motions, but typically require a predefined motion length, which is impractical in practical uses.
As a hybrid approach, BAMM~\cite{pinyoanuntapong2024bamm} combines both methods, generating coarse motions autoregressively and refining them bidirectionally. However, these models do not leverage part-based generation crucial for fine-grained control over individual body parts.

\paragraph{Part-Based Motion Generation}
Part-based approaches aim for detailed control over individual body parts. 
SCA~\cite{ghosh2021synthesis} trains separate networks for the upper and lower body but lacks coordination between them. 
AttT2M~\cite{zhong2023attt2m} uses body-part attention-based encoders to capture spatio-temporal features of motion but relies on a single decoder, limiting nuanced understanding of each part.

ParCo~\cite{zou2024parco} employs a VQ-VAE for each body part and shares token information among parts, enhancing global coherence while maintaining fine-grained control. However, excessive sharing can lead to overfitting and over-dependency among parts, limiting diversity and flexibility.
In bidirectional generation, this overfitting becomes more pronounced, causing conflicts in token alignment due to excessive interdependency. 

Our proposed model, BiPO, aims to combine the strengths of part-based and bidirectional approaches, such as enhanced control over individual body parts and comprehensive understanding of the extended sequences, while mitigating their limitations, including pronounced overfitting, the need for ground-truth motion length, and a lack of coordination over long temporal horizons.  

%% file: sec/3_approach.tex
\begin{figure*}[t]
  \centering
   \includegraphics[width=0.99\linewidth]{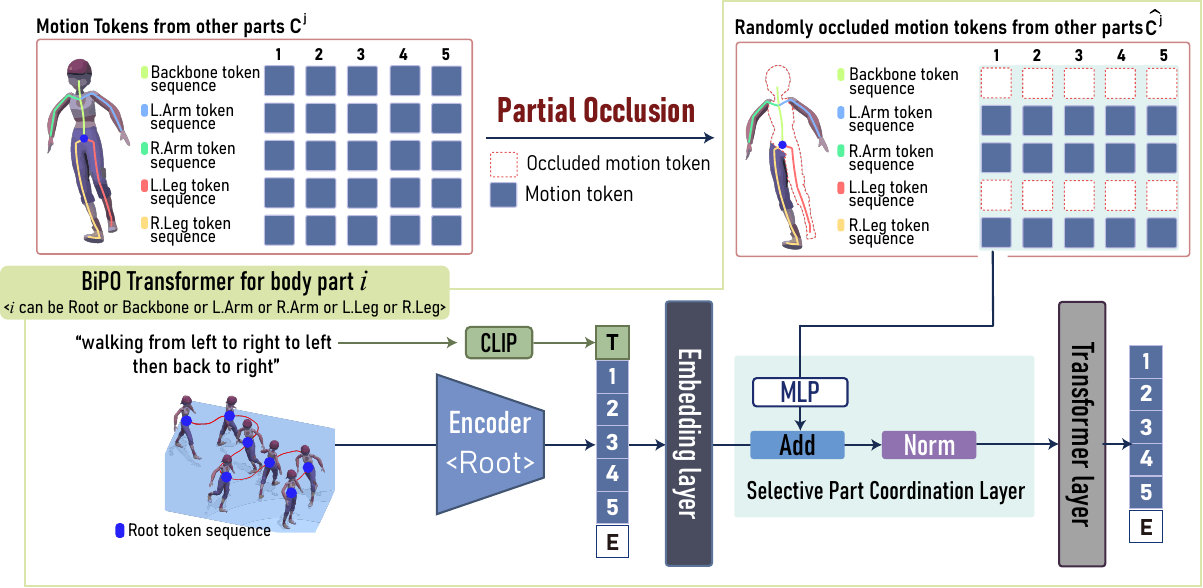}

   \caption{The architecture of our BiPO Transformer showing how the part-based learning is achieved. This is an example when \textit{i} is Root.}
   \label{fig:BiPO_Transformer}
\end{figure*}

\section{Approach}
\begin{figure*}[t]
  \centering
   \includegraphics[width=0.99\linewidth]{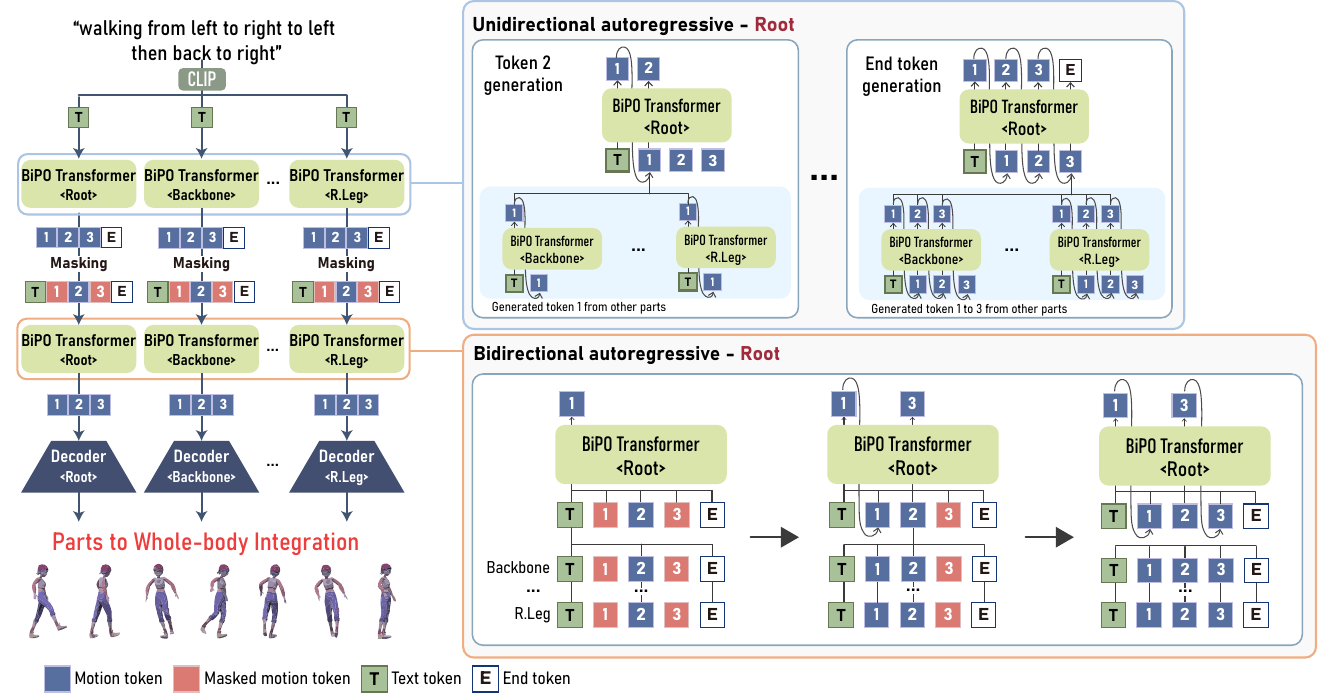}

   \caption{Dual-iteration Cascaded Part-based Motion Decoding. In the first iteration, each part undergoes autoregressive decoding with a unidirectional causal mask to generate coarse-grained motion and predict sequence length. In the second iteration, a bidirectional causal mask is applied, allowing part-based bidirectional decoding to remove and predict even indexed motion tokens, resulting in a refined and coordinated motion sequence across all parts.}
   \label{fig:inference}
\end{figure*}

BiPO is built upon a transformer-based architecture to capture the complex dependencies of individual body parts and the interactions between different parts during motion learning, as shown in~\autoref{fig:BiPO_Transformer}. It integrates both unidirectional and bidirectional processing when employing part-based motion generation, as shown in \autoref{fig:inference}. Specifically, each body part is generated separately to allow for fine-grained control, while coherence across the entire body is maintained by referencing motion tokens from other parts. To prevent excessive reliance on contextual information and preserve the independence of each part's motion, PO technique is applied.

\subsection{Part-based Bidirectional Autoregressive}
\label{sec:Part-based Bidirectional Autoregressive}

Our model introduces a part-based bidirectional autoregressive approach for text-to-motion synthesis, where each body part is independently modeled while leveraging bidirectional context from other parts to maintain overall motion coherence. 

We train a motion tokenizer following the methodology of ParCo~\cite{zou2024parco}, which quantizes continuous motion data into discrete tokens for effective modeling.
For each body part $i$, we obtain a sequence of quantized motion tokens \( c_{1:L}^i \), where $L$ denotes the length of the motion sequence, representing the number of frames. The input sequence for each part includes the shared text embedding \( c_0^i \) derived from a pre-trained CLIP model~\cite{radford2021learning}, the motion tokens \( c_{1:L}^i \), and an \texttt{[END]} token \( c_{L+1}^i \) indicating the end of the sequence. The text embedding \( c_0^i \) is common to all parts, as it encapsulates the overall textual description guiding the motion generation.

In the Transformer architecture, self-attention computes pairwise interactions between tokens, allowing the model to capture temporal dependencies and inter-part relationships. However, directly applying self-attention over all tokens can lead to excessive interdependency between body parts, causing them to become overly entangled and reducing the independence of individual motions.

To address this issue, we exploit a masking strategy, as used in BAMM~\cite{pinyoanuntapong2024bamm}, to control the flow of information in the attention mechanism. This regulates token interactions, balancing the need for contextual coherence with the preservation of part independence.

Specifically, we employ two types of attention masks during training: Unidirectional Causal Mask ($M_{uc}$) and Bidirectional Part-based Mask ($M_{bp}$). $M_{uc}$ enforces autoregressive generation, where each token can attend only to tokens at earlier positions. In this setting, only the text token $c_0^i$ is unmasked initially, and all motion tokens \( c_{1:L}^i \) and the \texttt{[END]} token \( c_{L+1}^i \) are masked. This ensures that the model generates the motion sequence step by step, conditioned solely on the text input and previous tokens.
On the other hand, $M_{bp}$ aims to leverage bidirectional context. In this setting, both the text token $c_0$ and the \texttt{[END]} token \( c_{L+1}^i \) are unmasked, along with a randomly selected subset of motion tokens \( c_{1:L}^i \) for each part. Masked tokens cannot attend to other masked tokens but can attend to unmasked tokens in both past and future positions across all parts. Unmasked tokens can attend to all other unmasked tokens.

The attention mechanism with the applied mask is defined as:

\begin{equation} 
\label{eq:attention_w_mask} 
\text{Attention} = \operatorname{Softmax} \left( \frac{Q K^\top}{\sqrt{d_k}} + M \right) V, 
\end{equation}
where \( Q \), \( K \), and \( V \) represent queries, keys, and values in the transformer architecture, and \( d_k \) denotes the dimensionality of the keys. The self-attention mask \( M \in \mathbb{R}^{(L+2) \times (L+2)} \) sets entries to zero for allowed attentions and negative infinity for disallowed ones, enforcing the mask during the softmax operation. 

The Bidirectional Part-based Mask \( M_{bp} \) is defined as follows:
\begin{equation} 
\label{eq:masked_attention} 
(M_{bp})_{qk} = \left\{ \begin{array}{ll} 0, & \text{if } (q \geq k \text{ and } q \notin U) \text{ or } (k \in U), \\
-\infty, & \text{otherwise}, \end{array} \right.
\end{equation}
where \( q, k \in [0, 1, 2, \ldots, L + 1] \) are indices of query and key tokens, and \( U = [u_0, u_1, \ldots] \) is the set of indices corresponding to unmasked motion tokens. The Unidirectional Causal Mask \( M_{uc} \) is a special case with \( U = \emptyset \).

Our training objective combines both unidirectional and bidirectional masking to reconstruct each part's motion sequence conditioned on the text embedding. We maximize the likelihood of motion token $l$ of body part $i$ under both masking schemes, denoted as \( p_\theta(c^i_l | M_{uc}) \) and \( p_\theta(c^i_l | M_{bp}) \), respectively. The training process is conducted by minimizing the sum of the loss functions for all body parts. The loss function for each body part is defined as follows:

\begin{equation}
\small
\begin{split}
\mathcal{L}^{i}_{\text{hybrid}} = -\mathbb{E}_{\mathbf{c} \sim p(\mathbf{c})} \Bigg[ & \lambda \sum_{l=1}^{L} \log p_\theta(c^i_l | M_{uc}) \\
&+ (1 - \lambda) \sum_{l=1}^{L} \log p_\theta(c^i_l | M_{bp}) \Bigg],
\end{split}
\end{equation}
where \( \lambda \) is a hyperparameter controlling the balance between unidirectional and bidirectional masking. An optimal performance is achieved when \( \lambda = 0.5 \).
In the bidirectional masking, we randomly mask between  50\% and 100\% of the motion tokens \( c_{1:L}^i \) in each part, achieving an optimal balance between leveraging bidirectional context and preserving the autoregressive nature of motion generation.

\subsection{Partial Occlusion}
\label{sec:PO}

In part-based motion generation models, a critical challenge is maintaining coherence between different body parts while preserving the independence of each part's motion. During training, the motion tokens generated for each body part must be coordinated to produce a cohesive whole-body motion—a process we refer to as part coordination.

To enhance this inter-part coordination and prevent excessive dependency among body parts, we introduce PO, which is a stochastic training technique that utilizes uniform random masking to selectively occlude motion tokens from specific body parts during training. Instead of consistently providing full motion data from all parts to each part's motion generator, PO randomly occludes certain tokens, supplying only partial information about other body parts. This selective masking encourages each part's motion generator to learn to produce coherent motions even when information from other parts is incomplete.

The implementation of PO occurs within the Selective Part Coordination Layer, as illustrated in ~\autoref{fig:BiPO_Transformer}. In this layer, a subset of tokens from other body parts is initially masked and remains masked throughout the sequence. By training under conditions of partial information, the model learns to synthesize coordinated movements despite incomplete inputs.

Consequently, the model becomes more robust during inference, adeptly handling incomplete or variable inputs without compromising coordination across body parts. This approach enhances the diversity, realism, and adaptability of the generated motions, leading to improvements in both the flexibility and quality of the outputs.

Formally, for the current target body part $i$, PO randomly masks a subset of input tokens from other parts $c^j$ (where $j \neq i$ and $j \in [1, ..., S]$, with $S$ representing the total number of body parts). 
Let $\hat{c^j}$ denote the motion tokens, including those masked tokens by PO. The coordination of the current part motion $c^i_{\text{coord}}$ can then be computed as follows:

\begin{equation}
c^i_{\text{coord}} = \text{LN}(c^i + \text{MLP}^i(\hat{c^j})),
\end{equation}
where LN denotes Layer Normalization. By combining $c^i$ with the transformed information from other parts through $\text{MLP}^i(\hat{c^j})$, the model captures complex dependencies across part motions while maintaining robustness to incomplete inputs. 

\begin{table*}[ht]
  \centering
  \begin{tabular}{lccccccccc}
    \toprule
    \multirow{2}{*}{Methods} & \multicolumn{3}{c}{R-Precision $\uparrow$} & \multirow{2}{*}{FID $\downarrow$} & \multirow{2}{*}{MM-Dist $\downarrow$} & \multirow{2}{*}{Diversity $\rightarrow$} & \multirow{2}{*}{MModality $\uparrow$} \\
    \cmidrule(lr){2-4}
            & Top-1 $\uparrow$ & Top-2 $\uparrow$ & Top-3 $\uparrow$ & & & & \\
    \midrule
    Real motion & $0.511^{\pm .003}$ & $0.703^{\pm .003}$ & $0.797^{\pm .002}$ & $0.002^{\pm .000}$ & $2.974^{\pm .008}$ & $9.503^{\pm .065}$ & - \\
    \midrule
    MDM\textsuperscript{§}~\cite{yuan2023physdiff} & $0.320^{\pm .005}$ & $0.498^{\pm .004}$ & $0.611^{\pm .007}$ & $0.544^{\pm .044}$ & $5.566^{\pm .027}$ & \underline{$9.559^{\pm .086}$} & \underline{$2.799^{\pm .072}$} \\
    MotionDiffuse\textsuperscript{§}~\cite{zhang2022motiondiffuse} & $0.491^{\pm .001}$ & $0.681^{\pm .001}$ & $0.782^{\pm .001}$ & $0.630^{\pm .011}$ & $3.113^{\pm .001}$ & $9.410^{\pm .049}$ & $1.553^{\pm .042}$ \\
    MLD\textsuperscript{§}~\cite{chen2023executing} & $0.481^{\pm .003}$ & $0.673^{\pm .003}$ & $0.772^{\pm .002}$ & $0.473^{\pm .013}$ & $3.196^{\pm .010}$ & $9.724^{\pm .082}$ & $2.413^{\pm .079}$ \\
    Fg-T2M\textsuperscript{§}~\cite{wang2023fg} & $0.492^{\pm .002}$ & $0.683^{\pm .003}$ & $0.783^{\pm .002}$ & $0.243^{\pm .019}$ & $3.109^{\pm .007}$ & $9.278^{\pm .072}$ & $1.614^{\pm .049}$ \\
    M2DM\textsuperscript{§}~\cite{kong2023priority} & $0.497^{\pm .003}$ & $0.682^{\pm .002}$ & $0.763^{\pm .003}$ & $0.352^{\pm .005}$ & $3.134^{\pm .010}$ & 9.926$^{\pm .073}$ & \textbf{3.587$^{\pm .072}$} \\
    ReMoDiffuse\textsuperscript{§}~\cite{zhang2023remodiffuse} & $0.510^{\pm .005}$ & $0.698^{\pm .006}$ & $0.795^{\pm .004}$ & $0.103^{\pm .004}$ & $2.974^{\pm .016}$ & $9.018^{\pm .075}$ & $1.795^{\pm .043}$ \\
    MoMask\textsuperscript{§}~\cite{guo2024momask} & $0.521^{\pm .002}$ & $0.713^{\pm .003}$ & $0.807^{\pm .002}$ & \underline{$0.045^{\pm .002}$} & $2.958^{\pm .008}$ & - & $1.241^{\pm .040}$ \\
    \midrule
    Text2Gesture~\cite{bhattacharya2021text2gestures} & $0.165^{\pm .001}$ & $0.267^{\pm .002}$ & $0.345^{\pm .002}$ & $7.664^{\pm .030}$ & $6.030^{\pm .008}$ & $6.409^{\pm .071}$ & - \\
    Seq2Seq~\cite{plappert2018learning} & $0.180^{\pm .002}$ & $0.300^{\pm .002}$ & $0.396^{\pm .002}$ & $11.75^{\pm .035}$ & $5.529^{\pm .007}$ & $6.223^{\pm .061}$ & - \\
    Language2Pose~\cite{ahuja2019language2pose} & $0.246^{\pm .001}$ & $0.387^{\pm .002}$ & $0.486^{\pm .002}$ & $11.02^{\pm .046}$ & $5.296^{\pm .008}$ & $7.676^{\pm .058}$ & - \\
    Hier~\cite{ghosh2021synthesis} & $0.301^{\pm .002}$ & $0.425^{\pm .002}$ & $0.552^{\pm .004}$ & $6.523^{\pm .024}$ & $5.012^{\pm .018}$ & $8.332^{\pm .042}$ & - \\
    TEMOS~\cite{petrovich2022temos} & $0.424^{\pm .002}$ & $0.612^{\pm .002}$ & $0.722^{\pm .002}$ & $3.734^{\pm .028}$ & $3.703^{\pm .008}$ & $8.973^{\pm .071}$ & $0.368^{\pm .018}$ \\
    TM2T~\cite{guo2022tm2t} & $0.424^{\pm .003}$ & $0.618^{\pm .003}$ & $0.729^{\pm .002}$ & $1.501^{\pm .017}$ & $3.467^{\pm .011}$ & $8.589^{\pm .076}$ & $2.424^{\pm .093}$ \\
    T2M~\cite{guo2022generating} & $0.455^{\pm .003}$ & $0.636^{\pm .003}$ & $0.736^{\pm .002}$ & $1.087^{\pm .021}$ & $3.347^{\pm .008}$ & $9.175^{\pm .083}$ & $2.219^{\pm .074}$ \\
    T2M-GPT~\cite{zhang2023generating} & $0.491^{\pm .003}$ & $0.680^{\pm .003}$ & $0.775^{\pm .002}$ & $0.116^{\pm .004}$ & $3.118^{\pm .011}$ & $9.761^{\pm .081}$ & $1.856^{\pm .011}$ \\
    AttT2M~\cite{zhong2023attt2m} & $0.499^{\pm .003}$ & $0.690^{\pm .002}$ & $0.786^{\pm .002}$ & $0.112^{\pm .006}$ & $3.038^{\pm .007}$ & $9.700^{\pm .090}$ & 2.452$^{\pm .051}$ \\
    ParCo~\cite{zou2024parco} & $0.515^{\pm .003}$ & $0.706^{\pm .003}$ & $0.801^{\pm .002}$ & $0.109^{\pm .005}$ & $2.927^{\pm .008}$ & $9.576^{\pm .088}$ & $1.382^{\pm .060}$ \\
    BAMM~\cite{pinyoanuntapong2024bamm} & \textbf{0.525$^{\pm .002}$} & \textbf{0.720$^{\pm .003}$} & \textbf{0.814$^{\pm .003}$} & $0.055^{\pm .002}$ & \underline{$2.919^{\pm .008}$} & $9.717^{\pm .089}$ & $1.687^{\pm .051}$ \\
    \midrule
    BiPO (Ours) & \underline{0.523$^{\pm .003}$} & \underline{$0.714^{\pm .002}$} & \underline{$0.809^{\pm .002}$} & \textbf{0.030$^{\pm .002}$} & \textbf{2.880$^{\pm .009}$} & \textbf{9.556$^{\pm .076}$} & $1.374^{\pm .047}$ \\
    \bottomrule
  \end{tabular}
  \caption{Comparative results on the HumanML3D test set against current state-of-the-art methods. Metrics where “↑” indicates that a higher value is preferred, “↓” indicates that a lower value is favorable, and “→” indicates metrics optimized when closer to real motion score of 9.503. The top result is highlighted in bold, with the second-best result underlined. The symbol § indicates evaluations performed using the ground-truth motion length.}
  \label{tab:humanml3d}
\end{table*}

\begin{figure*}[t]
  \centering
   \includegraphics[width=0.99\linewidth]{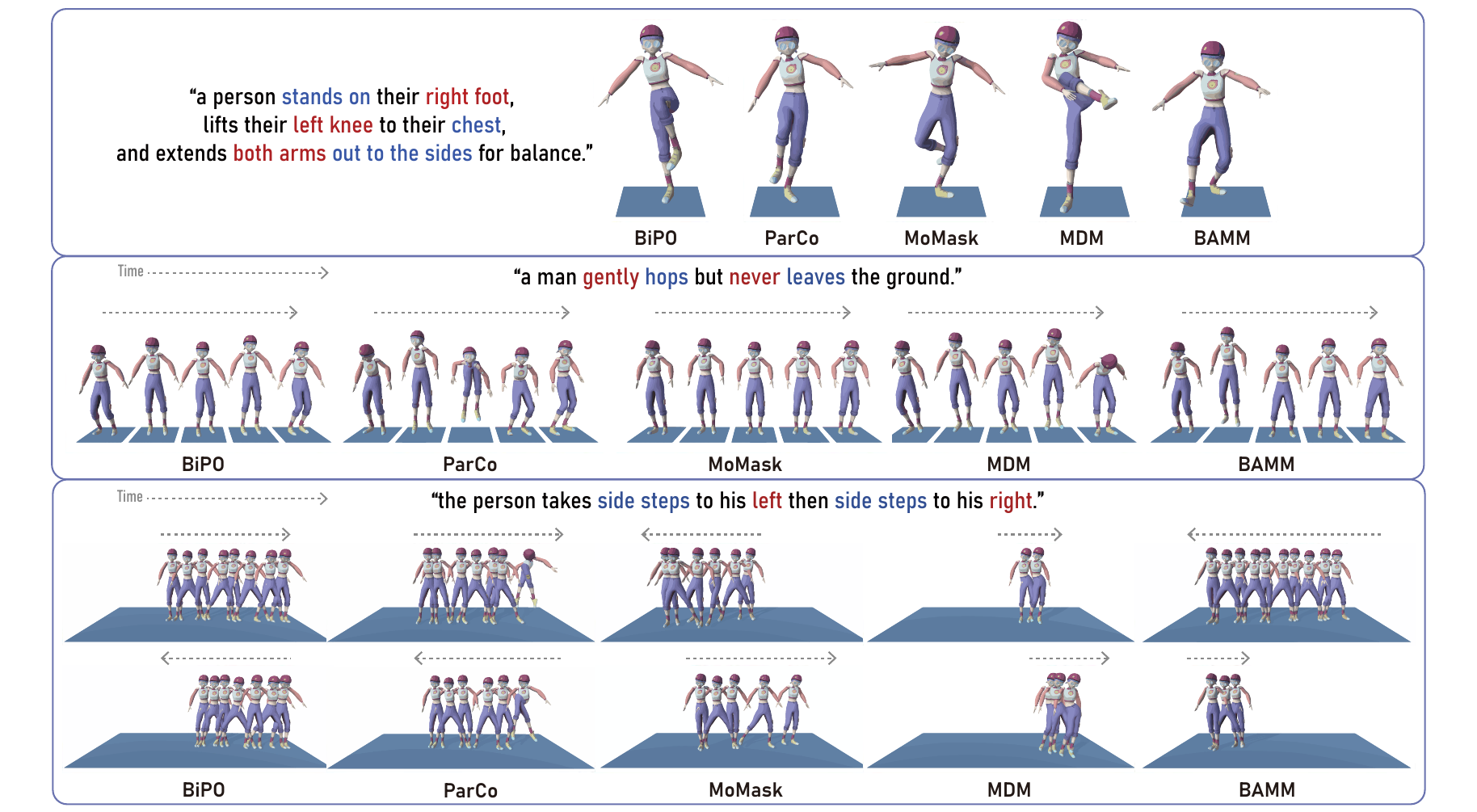}

   \caption{Qualitative comparison with existing methods. Words highlighted in red indicate the overall action, while words highlighted in blue specify how the action is performed.}
   \label{fig:Qualitative comparison}
\end{figure*}

\subsection{Inference}
\label{sec:Inference}

During inference, as illustrated in~\autoref{fig:inference}, our model generates the final motion sequence by utilizing the complete set of motion tokens from all body parts in an autoregressive manner. 

In the initial unidirectional generation phase, the model generates each body part's motion sequentially. At each time step $t$, for each part $i$, the model predicts the next token \( c_t^i \) by conditioning on its own past tokens \( c_1^i, \dots, c_{t-1}^i \) and the current tokens of all other parts \( \{c_t^j\} \) where \( j \neq i \). This dynamic conditioning allows the model to capture the interplay between different body parts, ensuring that the motions are synchronized and contextually correct.

Following the unidirectional generation, a bidirectional refinement step is involved to further enhance motion quality and inter-part coherence. In this phase, we selectively mask and regenerate tokens, specifically those at even indexed time steps, in the previously generated sequence using the bidirectional autoregressive approach. By incorporating both past and future contexts when predicting the masked tokens, the model can resolve any inconsistencies and fine-tune the motions to better align with the overall temporal dynamics and the textual description.

%% file: sec/4_experiments.tex
\section{Experiments}

\subsection{Experimental Setup}

We evaluate our proposed BiPO model on the widely used HumanML3D benchmark.
HumanML3D comprises 14,616 motion sequences collected from the AMASS~\cite{mahmood2019amass} and HumanAct12~\cite{guo2020action2motion} datasets, paired with 44,970 textual descriptions. 
Each motion is annotated with three different textual descriptions, covering a diverse range of actions including exercising, dancing, and acrobatics. The motions are represented using the SMPL skeletal model with 22 joints.
We adhere to the standard training, validation, and test splits with a ratio of 80:5:15. 
The dataset is augmented by mirroring during training to further increase the diversity of motions.
We adopt the pose representation from previous works such as T2M~\cite{guo2022generating}, and ensure that the dataset
are processed uniformly for training and evaluation.
The implementation details will be explained in the supplementary.

\subsection{Evaluation Metrics}
To assess the performance of our model, we employ several standard metrics commonly used in text-to-motion generation~\cite{guo2022generating}: R-precision, FID, MM-Dist, Diversity, and MModality. 

R-precision measures the semantic alignment between the generated motions and their corresponding textual descriptions. For each generated motion, we compute its similarity with the ground-truth text and compare it against 31 randomly selected mismatched texts. Higher R-precision indicates better text-motion alignment.

FID is the most crucial metric. It quantifies the distributional discrepancy between the generated motions and real motions in a feature space extracted by a pre-trained motion encoder. Lower FID values indicate that the generated motions are statistically closer to real human motions.

MM-Dist measures the Euclidean distance between the feature vectors of the generated motions and the corresponding textual descriptions. Lower MM-Dist values indicate better semantic similarity and alignment between text and motion.

Diversity evaluates the variance in generated motions. It is computed as the average pairwise Euclidean distance between randomly sampled motion features from the entire set of generated motions. A closer match in diversity between generated and real motions indicates greater alignment with real motion distributions.

MModality assesses the model's ability to generate diverse motions from the same textual description. For each text prompt, we generate motions 10 times and compute the average pairwise feature distance among them. Higher MModality scores suggest that the model can produce varied motions that all semantically align with the same text.

\subsection{Comparative Evaluation Results}
We conducted extensive experiments to compare our model with recent text-to-motion generation methods. 
We repeated each evaluation 20 times (5 times for MModality) and reported mean results with a 95\% confidence interval.
The quantitative results are presented in~\autoref{tab:humanml3d} and qualitative comparisons are shown in~\autoref{fig:Qualitative comparison}.

Our model outperforms previous methods in nearly all metrics. Notably, in the key metric FID, our model shows a substantially reduced FID score compared to SOTA models including MoMask, ParCo, and BAMM, reflecting an enhancement in the motion generation quality of BiPO.
Unlike ParCo, BiPO's bidirectional structure facilitates smooth transition and coordinated movements, reflecting a comprehensive understanding of overall motion patterns. Furthermore, unlike non-part-based models such as MoMask, MDM and BAMM, BiPO can accurately interpret specific textual prompts, such as `extends both arms' or leg-specific actions like `never leaves,' `hops,' and `side steps.' 

\subsection{Motion Editing Evaluation}
\label{sec:Motion Editing Evaluation}
We further evaluated our model's capabilities in motion editing tasks, with the results presented in~\autoref{tab:motion edit}. We conducted experiments on four motion editing scenarios to assess the model's adaptability and robustness: Temporal Inpainting, Temporal Outpainting, Prefix, and Suffix.
Temporal Inpainting involves filling in the middle 50\% of the motion sequence. Temporal Outpainting involves generating the outer 25\% at both the beginning and the end of the sequence, given the middle 50\%. 
Prefix involves generating the final 50\% of the sequence based on the initial 50\%, while Suffix involves generating the initial 50\% of the sequence based on the final 50\%.  Since motion editing aligns more closely with prediction than generation, we did not apply the MModality metric.

In all scenarios, our model outperforms existing methods, particularly in FID, indicating superior quality and realism in the generated motions. 

\subsection{Ablation Study}
\label{sec:Ablation Study}

We conducted an ablation study to analyze the effects of incorporating a bidirectional autoregressive (BA) approach and the PO technique. The baseline model employs only the part-based unidirectional autoregressive approach. 

As presented in~\autoref{tab:ablation}, integrating the BA into the baseline model significantly improves semantic alignment with the textual description and increases motion diversity. Furthermore, the inclusion of PO generally improves performance across all evaluation metrics. 
These results suggest that PO helps mitigate the over-reliance issues inherent in part-based BA methods, yielding more natural and coherent motions that closely reflect the input textual descriptions. 

\begin{table*}[ht]
  \centering
  \begin{tabular}{lccccccc}
    \toprule
    \multirow{2}{*}{Tasks} & \multirow{2}{*}{Methods} & \multicolumn{3}{c}{R-Precision $\uparrow$} & \multirow{2}{*}{FID $\downarrow$} & \multirow{2}{*}{MM-Dist $\downarrow$} & \multirow{2}{*}{Diversity $\rightarrow$} \\
    \cmidrule(lr){3-5}
    & & Top-1 $\uparrow$ & Top-2 $\uparrow$ & Top-3 $\uparrow$ & & \\
    \midrule
    Temporal Inpainting & MDM & 0.391 & 0.578 & 0.692 & 2.362 & 3.859 & 8.014 \\
    (In-betweening) & MoMask & 0.534 & 0.727 & 0.820 & 0.04 & 2.878 & 9.640 \\
    & BAMM & \textbf{0.535} & \textbf{0.729} & \textbf{0.821} & 0.056 & 2.863 & 9.629 \\
    & BiPO & 0.530 & 0.728 & \textbf{0.821} & \textbf{0.029} & \textbf{2.837} & \textbf{9.617} \\
    \midrule
    Temporal Outpainting & MDM & 0.415 & 0.613 & 0.727 & 2.057 & 3.619 & 8.199 \\
    & MoMask & 0.531 & 0.726 & 0.818 & 0.057 & 2.889 & \textbf{9.619} \\
    & BAMM & 0.535 & \textbf{0.730} & \textbf{0.822} & 0.056 & \textbf{2.856} & 9.659 \\
    & BiPO & \textbf{0.536} & 0.722 & 0.813 & \textbf{0.052} & 2.867 & 9.295 \\
    \midrule
    Prefix & MDM & 0.420 & 0.613 & 0.725 & 1.460 & 3.563 & 8.972 \\
    & MoMask & \textbf{0.536} & \textbf{0.730} & \textbf{0.823} & 0.060 & 2.875 & \textbf{9.607} \\
    & BAMM & 0.532 & 0.727 & 0.821 & 0.058 & \textbf{2.868} & 9.612 \\
    & BiPO & 0.522 & 0.717 & 0.808 & \textbf{0.036} & 2.876 & 9.357 \\
    \midrule
    Suffix & MDM & 0.403 & 0.597 & 0.711 & 2.562 & 3.731 & 8.088 \\
    & MoMask & 0.532 & \textbf{0.726} & \textbf{0.819} & 0.052 & 2.881 & 9.659 \\
    & BAMM & 0.527 & 0.720 & 0.814 & 0.050 & 2.891 & 9.721 \\
    & BiPO & \textbf{0.533} & 0.719 & 0.809 & \textbf{0.046} & \textbf{2.861} & \textbf{9.513} \\
    \bottomrule
  \end{tabular}
  \caption{Motion editing results. The target benchmark for Diversity is 9.503, corresponding to the Diversity of the real motion.}
  \label{tab:motion edit}
\end{table*}

\begin{table*}[ht]
  \centering
  \begin{tabular}{lccccccccc}
    \toprule
    \multirow{2}{*}{Methods} & \multicolumn{3}{c}{R-Precision $\uparrow$} & \multirow{2}{*}{FID $\downarrow$} & \multirow{2}{*}{MM-Dist $\downarrow$} & \multirow{2}{*}{Diversity $\rightarrow$} & \multirow{2}{*}{MModality $\uparrow$} \\
    \cmidrule(lr){2-4}
            & Top-1 $\uparrow$ & Top-2 $\uparrow$ & Top-3 $\uparrow$ & & & & \\
    \midrule
    baseline & $0.506^{\pm .003}$ & $0.701^{\pm .003}$ & $0.796^{\pm .002}$ & $0.108^{\pm .007}$ & $2.952^{\pm .008}$ & $9.544^{\pm .093}$ & 1.401$^{\pm .059}$ \\

    with BA & $0.517^{\pm .002}$ & $0.709^{\pm .003}$ & $0.804^{\pm .002}$ & $0.047^{\pm .002}$ & $2.915^{\pm .010}$ & \underline{9.497$^{\pm .081}$} & $1.278^{\pm .059}$ \\

    with PO (25\%) & $0.511^{\pm .003}$ & $0.704^{\pm .002}$ & $0.800^{\pm .002}$ & $0.065^{\pm .003}$ & $2.934^{\pm .009}$ & $9.563^{\pm .072}$ & $1.518^{\pm .115}$ \\

    with PO (30\%) & $0.507^{\pm .003}$ & $0.697^{\pm .002}$ & $0.793^{\pm .002}$ & $0.063^{\pm .004}$ & $2.967^{\pm .008}$ & 9.522$^{\pm .072}$ & $1.615^{\pm .075}$ \\

    with PO (35\%) & $0.507^{\pm .003}$ & $0.699^{\pm .002}$ & $0.795^{\pm .001}$ & $0.060^{\pm .004}$ & $2.962^{\pm .009}$ & $9.552^{\pm .081}$ & \textbf{1.674$^{\pm .061}$} \\

    with PO (40\%) & $0.505^{\pm .003}$ & $0.696^{\pm .003}$ & $0.794^{\pm .002}$ & $0.051^{\pm .004}$ & $2.969^{\pm .010}$ & $9.535^{\pm .077}$ & \underline{1.653$^{\pm .070}$} \\

    with BA + PO (25\%) & \underline{$0.520^{\pm .002}$} & $0.711^{\pm .002}$ & $0.807^{\pm .002}$ & $0.039^{\pm .003}$ & $2.889^{\pm .007}$ & \textbf{9.499$^{\pm .065}$} & $1.285^{\pm .054}$ \\

    with BA + PO (30\%) & $0.519^{\pm .003}$ & $0.713^{\pm .002}$ & \underline{$0.809^{\pm .002}$} & \underline{$0.032^{\pm .002}$} & $2.896^{\pm .009}$ & $9.449^{\pm .089}$ & $1.384^{\pm .010}$ \\

    with BA + PO (35\%) & \underline{0.520$^{\pm .003}$} & \textbf{0.715$^{\pm .002}$} & \textbf{0.810$^{\pm .002}$} & \underline{0.032$^{\pm .001}$} & \textbf{2.879$^{\pm .008}$} & 9.533$^{\pm .078}$ & 1.271$^{\pm .090}$ \\

    with BA + PO (40\%) & \textbf{0.523$^{\pm .003}$} & \underline{$0.714^{\pm .002}$} & \underline{$0.809^{\pm .002}$} & \textbf{0.030$^{\pm .002}$} & \underline{$2.880^{\pm .009}$} & $9.556^{\pm .076}$ & $1.374^{\pm .047}$ \\

    \bottomrule
  \end{tabular}
  \caption{Ablation Study. The percentage values next to PO indicate the likelihood of occluding each part's information within the PO technique. The target benchmark for Diversity is 9.503, corresponding to the Diversity of the real motion.}
  \label{tab:ablation}
\end{table*}

%% file: sec/5_user_study.tex
\begin{figure}[t]
  \centering
   \includegraphics[width=0.97\linewidth]{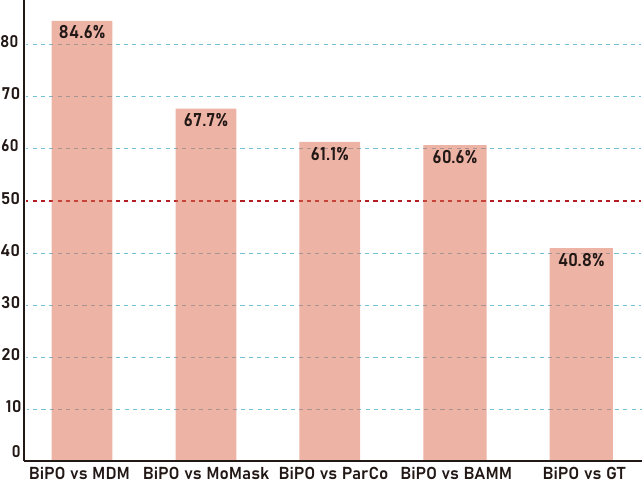}
   \caption{User study results showing BiPO's preference rate over other models, with a dashed red line at the 50\% threshold.}
   \label{fig:user study}
\end{figure}

\section{User Study}

Our user study results are shown in \autoref{fig:user study}. We conducted the study with 30 participants (19 males and 11 females, aged 20 to 45, $\mu=26.43$ and $ \sigma=5.68$). For each method-BiPO, MDM, MoMask, ParCo, and BAMM-as well as for ground-truth (GT) data, we generated 30 text-motion pairs using the same text input across the methods, while the GT data was sampled directly from the test dataset. Participants evaluated pairs of methods (e.g., BiPO vs. MoMask) in response to the question, ``\textit{Which motion better reflects the realism and alignment with the provided text description?}" The GT motion length was provided to MoMask and MDM for generation, but not for BiPO, ParCo, and BAMM. Overall, our model, BiPO, was preferred over the other models.

%% file: sec/6_conclusion.tex
\section{Conclusion}
In this paper, we have presented BiPO, a novel model that significantly advances the field of text-to-motion synthesis by effectively generating natural and expressive human motions from textual descriptions. The significance of our work lies in addressing the limitations of existing methods by combining the strengths of part-based and bidirectional approaches. By enhancing both global motion coherence and detailed control over individual body parts, BiPO produces highly natural and expressive motions that set a new benchmark in the field. 
We believe our work provides a foundation for future research in text-to-motion synthesis, leading to more intuitive and powerful tools for content creation. 

While BiPO achieves substantial progress, there remain promising directions for future exploration. Although PO mitigates interdependency among body parts, it currently relies on fixed masking probabilities. Investigating adaptive occlusion strategies that dynamically adjust occlusion probabilities based on context could further improve robustness. Furthermore, incorporating physics-based constraints could enhance the physical plausibility of generated motions, bringing them closer to real-world applicability.  

%% file: sec/X_suppl.tex
\clearpage
\setcounter{page}{1}
\maketitlesupplementary

\section{Overview}
The supplementary material is structured as follows:

\begin{itemize}
    \item Section 8: Implementation Details.
    \item Section 9: Body part division.
    \item Section 10: Reconstrcution of VQ-VAE.
    \item Section 11: Effect of Dual-iteration Cascaded Part-based Motion Decoding.
    \item Section 12: Feature extractor for evaluation.
    \item Section 13: Visualization for Motion editing.
    \item Section 14: Visualization for Ablation study.
    \item Section 15: Evaluation Metrics details.
    \item Section 16: Additional qualitative test.
    \item Section 17: Motion Representations.
    \item Section 18: Dataset Details.
    \item Section 19: User study details.
    \item Section 20: Experiments of KIT-ML.
    \item Section 21: Limitations.
    
\end{itemize}

\section{Implementation Details}
\subsection{Architecture}
Our method utilizes 6 lightweight VQ-VAEs~\cite{van2017neural} to discretize part motions and 6 small transformers~\cite{vaswani2017attention} equipped with Part Coordination modules for generating text-driven motions~\cite{zou2024parco}.
Each VQ-VAE contains a codebook with 512 entries. For most parts, the code dimension is set to 128, while the Root part has a reduced dimension of 64. The encoder applies a downsampling rate of \( r = 4 \) to reduce the motion sequence length.
The transformers consist of 14 layers, with each layer having a token dimension of 256. A Selective Part Coordination Layer is added before all remaining layers except the first transformer layer.
Selective Part Coordination Layer within the same layer of each transformer share their weights, and each Selective Part Coordination Layer includes 3 MLP layers.
For text-to-motion generation, we use the CLIP model~\cite{radford2021learning} with the ViT-B/32 variant to encode text features, enabling robust alignment between textual descriptions and motion representations.

\subsection{Hyperparameters}
During training, we employ a masking probability of 40\% within the Selective Part Coordination Layer to randomly mask out tokens from other parts, encouraging the model to generate coordinated motions even with incomplete information.
For training the VQ-VAE, we use a learning rate of \(2 \times 10^{-4}\) for the first 200K steps, and \(1 \times 10^{-5}\) after 100k steps.
The AdamW~\cite{loshchilov2017decoupled} optimizer is applied with \(\beta_1 = 0.9\) and \(\beta_2 = 0.99\), and the batch size is set to 256.
The commitment loss weight \(\beta\) is fixed at 1.0. For transformer training, we use a learning rate of \(1 \times 10^{-4}\) for the first 150K steps, decreasing to \(5 \times 10^{-6}\) after 150K.
The AdamW optimizer is also used for the transformer with \(\beta_1 = 0.5\) and \(\beta_2 = 0.99\), and the batch size is set to 64. All experiments are conducted using a single NVIDIA A6000 GPU and INTEL XEON(R) PLATINUM 8568Y+ in Ubuntu 22.04. The training of the VQ-VAE takes approximately 20 hours, while the training for text-to-motion generation takes around 64 hours.

\begin{figure}[t]
  \centering
   \includegraphics[width=0.99\linewidth]{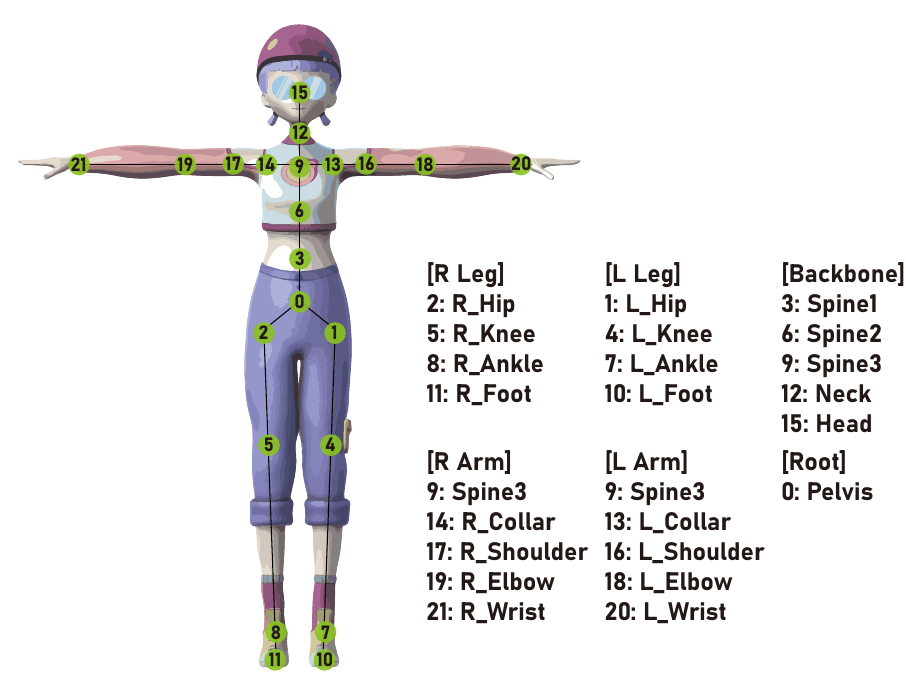}

   \caption{Visualization of body part division.}
   \label{fig:body partition}
\end{figure}

\section{Body Part Division}

Our method follows the same body partitioning strategy as ParCo~\cite{zou2024parco}, dividing the whole body into six parts: R.Leg, L.Leg, R.Arm, L.Arm, Backbone, and Root. The body partitioning is illustrated in \autoref{fig:body partition}. For our experiments, we exclusively used the HumanML3D dataset~\cite{guo2022generating}. Specifically, both R.Arm and L.Arm include the 9-th joint, as it serves as a critical key point connecting the arms to the backbone. This joint provides essential positional information for the arms relative to the connection point with the backbone.

During whole-body motion reconstruction from part motions, we generate three predictions for this joint: one from R.Arm, one from L.Arm, and one from Backbone. The final prediction for this joint is obtained by averaging these three values, ensuring consistent and accurate integration of part motions into the whole-body motion.

\begin{table*}[ht]
  \centering
  \begin{tabular}{lccccccccc}
    \toprule
    \multirow{2}{*}{Methods} & \multicolumn{3}{c}{R-Precision $\uparrow$} & \multirow{2}{*}{FID $\downarrow$} & \multirow{2}{*}{MM-Dist $\downarrow$} & \multirow{2}{*}{Diversity $\rightarrow$} & \multirow{2}{*}{MModality $\uparrow$} \\
    \cmidrule(lr){2-4}
            & Top-1 $\uparrow$ & Top-2 $\uparrow$ & Top-3 $\uparrow$ & & & & \\
    \midrule
    Real motion & $0.511^{\pm .003}$ & $0.703^{\pm .003}$ & $0.797^{\pm .002}$ & $0.002^{\pm .000}$ & $2.974^{\pm .008}$ & $9.503^{\pm .065}$ & - \\
    \midrule
    BiPO (R) & 0.500$^{\pm .003}$ & $0.690^{\pm .002}$ & $0.787^{\pm .002}$ & 0.020$^{\pm .000}$ & 3.026$^{\pm .006}$ & 9.430$^{\pm .094}$ & - \\
    BiPO (G) & 0.523$^{\pm .003}$ & $0.714^{\pm .002}$ & $0.809^{\pm .002}$ & 0.030$^{\pm .002}$ & 2.880$^{\pm .009}$ & 9.556$^{\pm .076}$ & $1.374^{\pm .047}$ \\
    \bottomrule
  \end{tabular}
  \caption{Reconstruction and Generation results. BiPO (R) represents the reconstruction performance of VQ-VAE, while BiPO (G) represents the performance of text-to-motion generation.}
  \label{tab:humanml3d recon}
\end{table*}

\begin{table*}[ht]
  \centering
  \begin{tabular}{lccccccccc}
    \toprule
    \multirow{2}{*}{Methods} & \multicolumn{3}{c}{R-Precision $\uparrow$} & \multirow{2}{*}{FID $\downarrow$} & \multirow{2}{*}{MM-Dist $\downarrow$} & \multirow{2}{*}{Diversity $\rightarrow$} & \multirow{2}{*}{MModality $\uparrow$} \\
    \cmidrule(lr){2-4}
            & Top-1 $\uparrow$ & Top-2 $\uparrow$ & Top-3 $\uparrow$ & & & & \\
    \midrule
    Real motion & $0.511^{\pm .003}$ & $0.703^{\pm .003}$ & $0.797^{\pm .002}$ & $0.002^{\pm .000}$ & $2.974^{\pm .008}$ & $9.503^{\pm .065}$ & - \\

    \midrule
    BiPO (- DC) & 0.519$^{\pm .003}$ & $0.712^{\pm .003}$ & $0.808^{\pm .002}$ & 0.034$^{\pm .002}$ & 2.895$^{\pm .009}$ & \textbf{9.496$^{\pm .076}$} & \textbf{1.406$^{\pm .052}$} \\
    BiPO & \textbf{0.523$^{\pm .003}$} & \textbf{0.714$^{\pm .002}$} & \textbf{0.809$^{\pm .002}$} & \textbf{0.030$^{\pm .002}$} & \textbf{2.880$^{\pm .009}$} & 9.556$^{\pm .076}$ & $1.374^{\pm .047}$ \\
    \bottomrule
  \end{tabular}
  \caption{Ablation for Dual-iteration Cascaded Part-based Motion Decoding. BiPO (- DC) represents the results generated without employing Dual-iteration Cascaded Part-based Motion Decoding. Better result is highlighted in bold.}
  \label{tab:humanml3d Dual-iteration Cascaded Part-based Motion Decoding}
\end{table*}

\begin{figure}[t]
  \centering
   \includegraphics[width=0.99\linewidth]{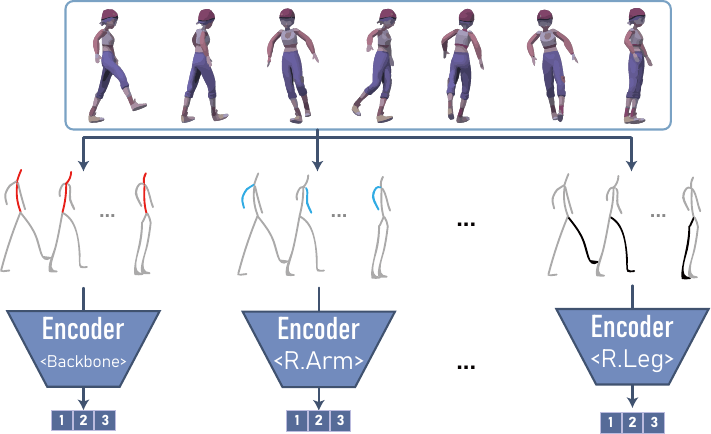}

   \caption{Part-based VQ-VAE architecture.}
   \label{fig:VQVAE}
\end{figure}

\section{Reconstruction of VQ-VAE}
We utilized part-based VQ-VAEs as proposed by ParCo~\cite{zou2024parco}, illustrated in \autoref{fig:VQVAE}.
The reconstruction performance is presented in \autoref{tab:humanml3d recon}.
For evaluation, the reconstructed motions are integrated into whole-body motion sequences.
While ParCo employs the VQ-VAE trained at the final iteration during training, we selected the VQ-VAE that achieved the best FID performance on the validation dataset.
Experimentally, ParCo demonstrated the highest text-to-motion performance using the VQ-VAE from the final iteration.
However, in our model, BiPO, selecting the VQ-VAE with the best FID performance on the validation dataset yield better overall performance, leading us to adopt this approach.

\section{Effect of Dual-iteration Cascaded Part-based Motion Decoding}
We conduct performance evaluations on the effectiveness of Dual-iteration Cascaded Part-based Motion Decoding. Dual-iteration Cascaded Motion Decoding, proposed by BAMM~\cite{pinyoanuntapong2024bamm}, is reported to outperform other masking strategies by masking even indexed motion tokens and predicting them again. Based on this, our model, BiPO, also adopt this masking strategy. Additionally, we conduct further experiments to evaluate the effectiveness of this approach. The performance results, present in \autoref{tab:humanml3d Dual-iteration Cascaded Part-based Motion Decoding}, demonstrate that Dual-iteration Cascaded Part-based Motion Decoding is indeed effective.

\section{Feature extractor for evaluation}

For evaluation, we utilize a motion feature extractor and a text feature extractor trained using contrastive learning introduced by T2M~\cite{guo2022generating}. These extractors are specifically designed to map text and motion features into a shared embedding space, where matching pairs are positioned closer together, and non-matching pairs are separated. This approach enables effective alignment of text-motion pairs for accurate evaluation. By employing the contrastive learning-based feature extractor, our evaluation framework aligns with established benchmarks, ensuring a rigorous assessment of text-to-motion alignment and generation quality.

\section{Visualization for Motion editing}
\begin{figure*}[t]
  \centering
   \includegraphics[width=0.99\linewidth]{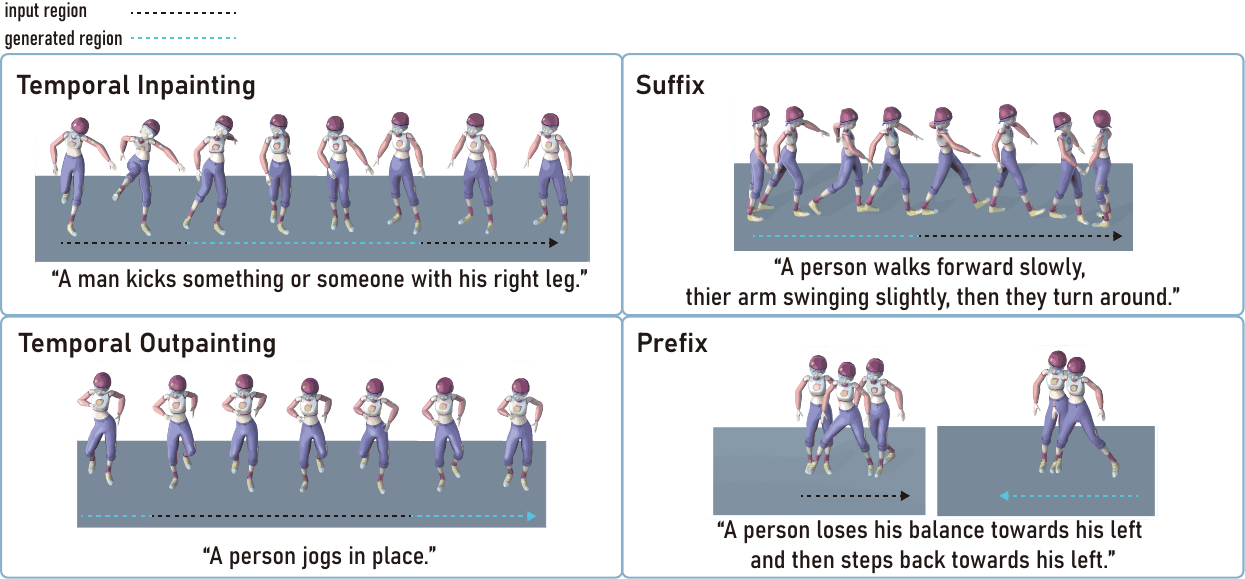}

   \caption{Visualization for Motion editing. The input region represents the real motion used as a condition, while the generated region refers to the motion generated by Our model, BiPO.}
   \label{fig:Visualization for Motion editing}
\end{figure*}

We visualize motion editing results. Visualizations for Motion editing are illustrated in \autoref{fig:Visualization for Motion editing}. In all four cases (Temporal Inpaintin, Temporal Outpainting, Prefix and Suffix), the remaining motions are appropriately generated to align with the given condition, demonstrating the effectiveness of our proposed model, BiPO, in performing the motion editing task.

\section{Visualization for Ablation study}
\begin{figure*}[ht]
    \centering
     \includegraphics[width=0.99\linewidth]{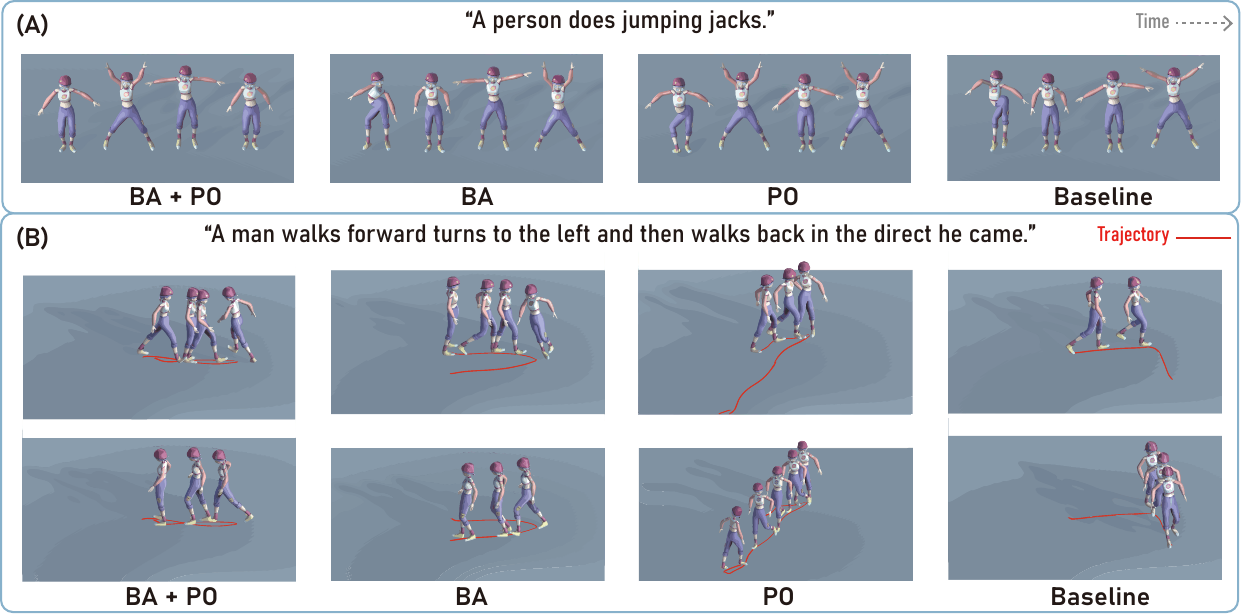}
    
    \caption{Visualization for Ablation study. (A) describes a motion performed in place, such as "A person does jumping jacks." (B) describes a motion involving walking, such as "A man walks forward, turns to the left, and then walks back in the direction he came from."}
    \label{fig:Visualization for Ablation study}
\end{figure*}

We visualize an ablation study.
Visualizations for Motion editing are illustrated in \autoref{fig:Visualization for Ablation study}. This ablation study demonstrates the effectiveness of the proposed part-based bidirectional autoregressive approach and Partial Occlusion.

\section{Evaluation Metrics details}
We use several evaluation metrics, as proposed in T2M ~\cite{guo2022generating}, to measure the performance of our model. Below, we provide detailed formulations for these metrics.

\subsection{Fréchet Inception Distance}
Fréchet Inception Distance (FID) evaluates the quality of the generated motions by comparing the distribution of their features to the distribution of ground-truth motion features. It is calculated as follows:
\begin{equation}
\text{FID} = \|\mu_\text{gt} - \mu_\text{pred}\|^2 + \mathrm{Tr}(\Sigma_\text{gt} + \Sigma_\text{pred} - 2(\Sigma_\text{gt} \Sigma_\text{pred})^{1/2}),
\end{equation}
where \(\mu_\text{gt}\) and \(\mu_\text{pred}\) are the mean feature vectors of the ground-truth and predicted motions, respectively. \(\Sigma_\text{gt}\) and \(\Sigma_\text{pred}\) represent their covariance matrices, and \(\mathrm{Tr}\) denotes the trace of a matrix.

\subsection{R-Precision}
R-Precision evaluates the semantic alignment between text descriptions and generated motions by measuring the retrieval accuracy of the most relevant matches. Each text description's corresponding motion feature is expected to appear within the top \( k \) nearest neighbors of the motion features retrieved from the generated data. 

To compute R-Precision, let \( f_\text{pred} \) and \( f_\text{text} \) represent the generated motion and the features of the text description, respectively. The distance matrix between all pairs of \( f_\text{pred} \) and \( f_\text{text} \) is defined as:
\begin{equation}
\mathbf{D}(i, j) = \|f_{\text{pred}, i} - f_{\text{text}, j}\|,
\end{equation}
where \( \mathbf{D}(i, j) \) denotes the Euclidean distance between the \( i \)-th generated motion feature and the \( j \)-th text feature.

For a given generated motion feature, 
we randomly sample 31 text descriptions from the test dataset.
Along with the text description \( f_{\text{text}, i}\) matched to \( f_{\text{pred}, i}\), these 32 text descriptions form the search pool. The top \( k \) nearest text descriptions are retrieved by sorting the distances in ascending order. The R-Precision at top-\( k \) is calculated as:
\begin{equation}
\text{R-Precision@k} = \frac{1}{N} \sum_{i=1}^{N} \mathbb{1} \{i \in \text{Top-}k(\mathbf{D}(i, :))\},
\end{equation}
where \( N \) is the total number of text-motion pairs, \( \mathbb{1} \{ \cdot \} \) is the indicator function that equals 1 if the condition is true and 0 otherwise, and \( \text{Top-}k(\mathbf{D}(i, :)) \) represents the indices of the top-\( k \) closest text descriptions from the search pool to the \( i \)-th generated motion feature.

In our experiments, we report R-Precision for \( k = 1, 2, \) and \( 3 \) to provide a comprehensive evaluation of the alignment between text and motion.

\subsection{MultiModal Distance}
MultiModal Distance (MM-Dist) measures the semantic alignment between the textual descriptions and the generated motions. It is defined as:
\begin{equation}
\text{MM-Dist} = \frac{1}{N} \sum_{i=1}^{N} \|f_{\text{pred},i} - f_{\text{text},i}\|,
\end{equation}
where \(f_{\text{pred},i}\) and \(f_{\text{text},i}\) are the features of the \(i\)-th generated motion and its corresponding text description, respectively.

\subsection{Diversity}
Diversity measures the variance of the generated motion sequences. For \(S_\text{dis}\) randomly sampled motion pairs, the diversity is computed as:
\begin{equation}
\text{Diversity} = \frac{1}{S_\text{dis}} \sum_{i=1}^{S_\text{dis}} \|f_{\text{pred},i} - f'_{\text{pred},i}\|,
\end{equation}
where \(f_{\text{pred},i}\) and \(f'_{\text{pred},i}\) are the features of the \(i\)-th pair of generated motions. In our experiments, \(S_\text{dis}\) is set to 300, following T2M ~\cite{guo2022generating}.

\subsection{Multimodality}
Multimodality (MModality) evaluates the diversity of motions generated from the same text description. For the \(r\)-th text prompt, we generate 30 motions and randomly sample two subsets, each containing 10 motions. The metric is calculated as:
\begin{equation}
\text{MModality} = \frac{1}{N} \sum_{i=1}^{N} \frac{1}{10} \sum_{j=1}^{10} \|f_{\text{pred},i,j} - f'_{\text{pred},i,j}\|,
\end{equation}
where \(f_{\text{pred},i,j}\) and \(f'_{\text{pred},i,j}\) are the features of the \(i\)-th pair of generated motions for the \(r\)-th text description.

\begin{figure*}[ht]
    \centering
        \includegraphics[width=0.99\linewidth]{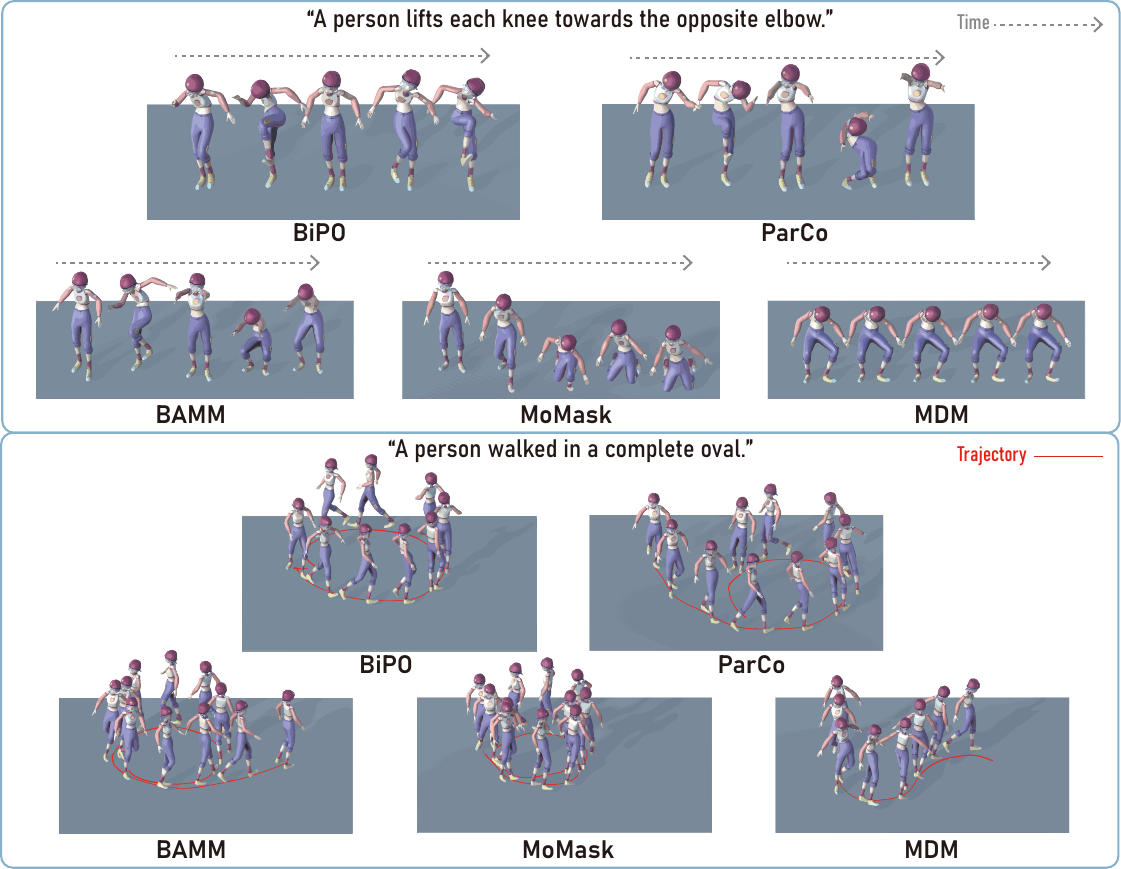}
    \caption{Additional qualitative test.}
    \label{fig:Additional qualitative test}
\end{figure*}

\section{Additional qualitative test}
We visualize additional qualitative tests. The examples are shown in \autoref{fig:Additional qualitative test}, featuring motions generated based on text prompts from the HumanML3D test set. These examples highlight the capability of our method to produce natural and well-coordinated motions that correspond closely to the input text descriptions.

\section{Motion Representations}
For motion representation, we follow T2M ~\cite{guo2022generating}. Each pose is described by:
\begin{equation}
(\dot{r}_a, \dot{r}_x, \dot{r}_z, r_y, j_p, j_v, j_r, c_f),
\end{equation}
where \(\dot{r}_a\) is the global root angular velocity; \(\dot{r}_x, \dot{r}_z\) are the root velocities in the X-Z plane; \(j_p, j_v, j_r\) represent joint positions, velocities, and rotations, respectively; and \(c_f\) denotes foot contact features derived from the heel and toe joint velocities. Each pose is represented as a feature vector with a total dimension of 263.

\section{Dataset Details}
The HumanML3D dataset~\cite{guo2022generating} is constructed by combining motion sequences from two large-scale publicly available datasets, HumanAct12~\cite{guo2020action2motion} and AMASS~\cite{mahmood2019amass}. These datasets consist of various types of human actions, including everyday activities such as walking and jumping, sports like swimming and karate, acrobatic movements such as cartwheels, and artistic performances like dancing.

The dataset is processed to ensure consistency and usability. Motion sequences are normalized to 20 frames per second (FPS), and those exceeding 10 seconds in duration are randomly cropped to 10 seconds. Each motion sequence is retargeted to a standardized human skeletal template and oriented to initially face the positive Z-axis.

To provide textual descriptions for the motions, annotations are collected through Amazon Mechanical Turk (AMT). Annotators are required to describe each motion with at least five words, and three descriptions are provided for each motion clip by different individuals. These descriptions undergo a post-processing step to remove inconsistencies or errors, resulting in high-quality textual annotations.

The final HumanML3D dataset comprises 14,616 motion sequences with a total of 44,970 textual descriptions, featuring a vocabulary of 5,371 unique words. The dataset spans approximately 28.59 hours of motion data, with an average clip length of 7.1 seconds, ranging from 2 to 10 seconds. The average textual description length is 12 words, with a median of 10 words. This makes HumanML3D one of the most extensive datasets for research involving text-to-motion synthesis. The dataset was further augmented using mirroring techniques to increase diversity. For example, a motion described as ``A man kicks something or someone with his left leg" was mirrored to create a new motion with the description ``A man kicks something or someone with his right leg." This approach ensures a balanced representation of left and right directional movements in the dataset.

\begin{table*}[ht]
  \centering
  \begin{tabular}{lccccccccc}
    \toprule
    \multirow{2}{*}{Methods} & \multicolumn{3}{c}{R-Precision $\uparrow$} & \multirow{2}{*}{FID $\downarrow$} & \multirow{2}{*}{MM-Dist $\downarrow$} & \multirow{2}{*}{Diversity $\rightarrow$} & \multirow{2}{*}{MModality $\uparrow$} \\
    \cmidrule(lr){2-4}
            & Top-1 $\uparrow$ & Top-2 $\uparrow$ & Top-3 $\uparrow$ & & & & \\
    \midrule
    Real motion & $0.424^{\pm .005}$ & $0.649^{\pm .006}$ & $0.779^{\pm .006}$ & $0.031^{\pm .004}$ & $2.788^{\pm .012}$ & $11.08^{\pm .097}$ & - \\
    \midrule
    MDM\textsuperscript{§}~\cite{yuan2023physdiff} & $0.164^{\pm .004}$ & $0.291^{\pm .004}$ & $0.396^{\pm .004}$ & $0.497^{\pm .021}$ & $9.191^{\pm .022}$ & $10.85^{\pm .109}$ & $1.907^{\pm .214}$ \\
    MotionDiffuse\textsuperscript{§}~\cite{zhang2022motiondiffuse} & $0.417^{\pm .004}$ & $0.621^{\pm .004}$ & $0.739^{\pm .004}$ & $1.954^{\pm .062}$ & $2.958^{\pm .005}$ & \textbf{11.10$^{\pm .143}$} & $0.730^{\pm .013}$ \\
    MLD\textsuperscript{§}~\cite{chen2023executing} & $0.390^{\pm .008}$ & $0.609^{\pm .008}$ & $0.734^{\pm .007}$ & $0.404^{\pm .027}$ & $3.204^{\pm .027}$ & $10.80^{\pm .117}$ & $2.192^{\pm .071}$ \\
    Fg-T2M\textsuperscript{§}~\cite{wang2023fg} & $0.418^{\pm .005}$ & $0.626^{\pm .004}$ & $0.745^{\pm .004}$ & $0.571^{\pm .047}$ & $3.114^{\pm .015}$ & $10.93^{\pm .083}$ & $1.019^{\pm .029}$ \\
    M2DM\textsuperscript{§}~\cite{kong2023priority} & $0.416^{\pm .004}$ & $0.628^{\pm .004}$ & $0.743^{\pm .004}$ & $0.515^{\pm .029}$ & $3.015^{\pm .017}$ & 11.417$^{\pm .97}$ & \textbf{3.325$^{\pm .37}$} \\
    ReMoDiffuse\textsuperscript{§}~\cite{zhang2023remodiffuse} & $0.427^{\pm .014}$ & $0.641^{\pm .004}$ & $0.765^{\pm .055}$ & \textbf{0.155$^{\pm .006}$} & $2.814^{\pm .012}$ & $10.80^{\pm .105}$ & $1.239^{\pm .028}$ \\
    MoMask\textsuperscript{§}~\cite{guo2024momask} & $0.433^{\pm .007}$ & $0.656^{\pm .005}$ & $0.781^{\pm .005}$ & $0.204^{\pm .011}$ & $2.779^{\pm .022}$ & - & $1.131^{\pm .043}$ \\
    \midrule
    Text2Gesture~\cite{bhattacharya2021text2gestures} & $0.156^{\pm .004}$ & $0.255^{\pm .004}$ & $0.338^{\pm .005}$ & $12.12^{\pm .183}$ & $6.946^{\pm .029}$ & $9.334^{\pm .079}$ & - \\
    Seq2Seq~\cite{plappert2018learning} & $0.103^{\pm .003}$ & $0.178^{\pm .005}$ & $0.241^{\pm .006}$ & $24.86^{\pm .348}$ & $7.960^{\pm .031}$ & $6.744^{\pm .106}$ & - \\
    Language2Pose~\cite{ahuja2019language2pose} & $0.221^{\pm .005}$ & $0.373^{\pm .004}$ & $0.483^{\pm .005}$ & $6.545^{\pm .072}$ & $5.147^{\pm .030}$ & $9.073^{\pm .100}$ & - \\
    Hier~\cite{ghosh2021synthesis} & $0.255^{\pm .006}$ & $0.432^{\pm .007}$ & $0.531^{\pm .007}$ & $5.203^{\pm .107}$ & $4.986^{\pm .027}$ & $9.563^{\pm .072}$ & $2.090^{\pm .083}$ \\
    TEMOS~\cite{petrovich2022temos} & $0.353^{\pm .006}$ & $0.561^{\pm .007}$ & $0.687^{\pm .005}$ & $3.717^{\pm .051}$ & $3.417^{\pm .019}$ & $10.84^{\pm .100}$ & $0.532^{\pm .034}$ \\
    TM2T~\cite{guo2022tm2t} & $0.280^{\pm .005}$ & $0.463^{\pm .006}$ & $0.587^{\pm .005}$ & $3.599^{\pm .153}$ & $4.591^{\pm .026}$ & $9.473^{\pm .117}$ & \underline{$3.292^{\pm .081}$} \\
    T2M~\cite{guo2022generating} & $0.361^{\pm .006}$ & $0.559^{\pm .007}$ & $0.681^{\pm .007}$ & $3.022^{\pm .107}$ & $3.488^{\pm .028}$ & $10.72^{\pm .145}$ & $2.052^{\pm .107}$ \\
    T2M-GPT~\cite{zhang2023generating} & $0.402^{\pm .006}$ & $0.619^{\pm .005}$ & $0.737^{\pm .006}$ & $0.717^{\pm .041}$ & $3.053^{\pm .026}$ & $10.86^{\pm .094}$ & $1.912^{\pm .036}$ \\
    AttT2M~\cite{zhong2023attt2m} & $0.413^{\pm .006}$ & $0.632^{\pm .006}$ & $0.751^{\pm .006}$ & $0.870^{\pm .039}$ & $3.039^{\pm .021}$ & $10.96^{\pm .123}$ & 2.281$^{\pm .047}$ \\
    ParCo~\cite{zou2024parco} & $0.430^{\pm .004}$ & $0.649^{\pm .007}$ & $0.772^{\pm .006}$ & $0.453^{\pm .027}$ & $2.820^{\pm .028}$ & $10.95^{\pm .094}$ & $1.245^{\pm .022}$ \\
    BAMM~\cite{pinyoanuntapong2024bamm} & \underline{0.438$^{\pm .009}$} & \underline{0.661$^{\pm .009}$} & \underline{0.788$^{\pm .005}$} & 0.183$^{\pm .013}$ & \underline{$2.723^{\pm .026}$} & \underline{$11.008^{\pm .094}$} & $1.609^{\pm .065}$ \\
    \midrule
    BiPO (Ours) & \textbf{0.444$^{\pm .005}$} & \textbf{0.674$^{\pm .006}$} & \textbf{0.803$^{\pm .005}$} & \underline{0.164$^{\pm .008}$} & \textbf{2.658$^{\pm .015}$} & 10.833$^{\pm .111}$ & $1.098^{\pm .047}$ \\
    \bottomrule
  \end{tabular}
  \caption{Comparative results on the KIT-ML test set against current state-of-the-art methods. Metrics where “↑” indicates that a higher value is preferred, “↓” indicates that a lower value is favorable, and “→” indicates metrics optimized when closer to real motion score of 11.08. The top result is highlighted in bold, with the second-best result underlined. The symbol § indicates evaluations performed using the ground-truth motion length.}
  \label{tab:kit}
\end{table*}

\begin{figure*}[t]
  \centering
   \includegraphics[width=0.99\linewidth]{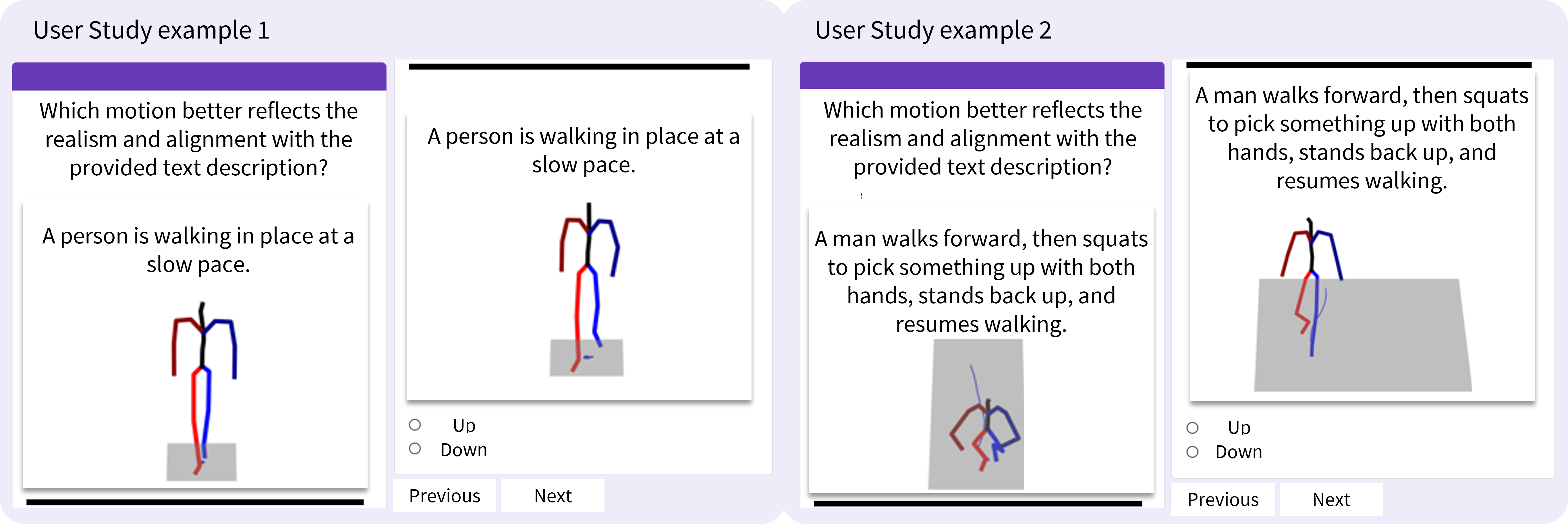}

   \caption{Examples of user study.}
   \label{fig:user study example}
\end{figure*}

\section{User study details}
For the user study, we utilize Google Forms. Examples of the survey are shown in \autoref{fig:user study example}. We sample 30 motions generated from the same textual prompts in the test dataset.
To ensure fairness, the models are anonymized, and their order is randomized for each question, with one model's motion displayed above and the other's below.
This approach prevented users from knowing which motion corresponded to which model, allowing for an unbiased evaluation.

\section{Experiments of KIT-ML}
We conducted further experiments on KIT-ML, a well-known dataset in text-to-motion synthesis. The results are presented in the ~\autoref{tab:kit}. As shown, our model, BiPO, achieves state-of-the-art performance, demonstrating its strong generalization capability.

\section{Limitations}
A limitation of BiPO lies in its inference speed. As an autoregressive model, BiPO inherently relies on sequential generation, which is slower compared to non-autoregressive methods. This limitation is further compounded by the model's part-based generation process, where each body part is generated individually. 

Additionally, the masking strategy employed by BiPO requires Dual-iteration Cascaded Part-based Motion Decoding. This repeated application of autoregressive generation across body parts and masked regions results in computational overhead, making the overall process slower.

% However, despite these limitations, BiPO is still significantly faster than the MDM~\cite{tevet2023human} model.
% On an NVIDIA A6000 GPU, our model achieves an inference time of approximately 1.348 seconds per sample, demonstrating its practical efficiency compared to MDM.

Although BiPO demonstrates slower inference speed compared to non-autoregressive methods, it achieves high-quality results.
% , outperforming the MDM model in terms of inference time.
Notably, even without employing the Dual-iteration Cascaded Part-based Motion Decoding, BiPO achieves state-of-the-art performance in FID. Skipping this step could make BiPO even faster, offering a speed advantage while maintaining competitive motion quality.